\documentclass[conference]{IEEEtran}
\IEEEoverridecommandlockouts
\usepackage{cite}
\usepackage{amsmath,amssymb,amsfonts,color}
\usepackage{algorithmic}
\usepackage{graphicx}
\usepackage{textcomp,bm}
\usepackage{xcolor}
\usepackage{hyperref} 
\usepackage{multirow}
\usepackage{threeparttable}

\allowdisplaybreaks

\begin{document}
	
\title{%Efficient and Robust Time-optimal Quadrotor Flight\\
%	Time-Adaptive Control for Rapid and Accurate Trajectory Tracking in Agile Quadrotor Flight
Efficient and Robust Time-Optimal Trajectory Planning and Control for Agile Quadrotor Flight
}

\author{Ziyu Zhou,  Gang Wang, %~\IEEEmembership{Senior Member,~IEEE}, 
	Jian Sun,~\IEEEmembership{Senior Member,~IEEE}, Jikai Wang,
		and	Jie Chen,~\IEEEmembership{Fellow,~IEEE}
	\thanks{The work was partially supported by the National Natural Science Foundation of China under Grants 62173034, 61925303, and 62088101.}
	\thanks{	Ziyu Zhou, Gang Wang, Jian Sun, and Jikai Wang are with the National Key Lab of Autonomous Intelligent Unmanned Systems, 
		Beijing Institute of Technology, Beijing 100081, China, and the Beijing Institute of Technology Chongqing Innovation Center, Chongqing 401120, China (e-mail: ziyuzhou@bit.edu.cn, gangwang@bit.edu.cn; sunjian@bit.edu.cn; jikaiwang@bit.edu.cn). Jie Chen is with the National Key Lab of Autonomous Intelligent Unmanned Systems,  Tongji University, Shanghai 201804, China (e-mail: chenjie@bit.edu.cn).}
	
}

%\and

\maketitle

\begin{abstract}

Agile quadrotor flight relies on rapidly planning and accurately tracking time-optimal trajectories, a technology critical to their application in the wild. However, the computational burden of computing time-optimal trajectories based on the full quadrotor dynamics (typically on the order of minutes or even hours) can hinder its ability to respond quickly to changing scenarios. Additionally, modeling errors and external disturbances can lead to deviations from the desired trajectory during tracking in real time. This letter proposes a novel approach to computing time-optimal trajectories, by fixing the nodes with waypoint constraints and adopting separate sampling intervals for trajectories between waypoints, which significantly accelerates trajectory planning. Furthermore, the planned paths are tracked via a time-adaptive model predictive control scheme whose allocated tracking time can be adaptively adjusted on-the-fly, therefore enhancing the tracking accuracy and robustness. We evaluate our approach through simulations and experimentally validate its performance in dynamic waypoint scenarios for time-optimal trajectory replanning and trajectory tracking.

\end{abstract}

\section*{Supplementary Material}
{\bf Code:} \url{https://github.com/BIT-KAUIS/Fast-fly}

{\bf Video:} \url{https://youtu.be/E6QVHWcvB6E}

\section{Introduction}

The use of quadrotors in both academic and commercial fields has garnered significant attention due to their excellent maneuverability and versatility  \cite{loianno2020special,chen2022unmanned}. Quadrotors have numerous applications, including aerial photography and videography, search and rescue operations, agricultural monitoring, package delivery, and military reconnaissance, to just name a few. These applications require the quadrotor to visit multiple waypoints within a limited flight duration due to on-board battery capacity, which necessitates time-optimal flight to improve the efficiency and task completion  \cite{hanover2023autonomous}.

Trajectory planning for time-optimal flight has always been a challenging problem \cite{foehn2021time}. Unlike the point-mass model, which has a closed-form solution with bang-bang acceleration and can quickly obtain the time-optimal trajectory by sampling at waypoints \cite{romero2022time}, the quadrotor model is underactuated. The thrust generated by the propellers can only act along the $z$-axis of the body, and both the thrust magnitude and direction need to be adjusted simultaneously to control the movement of the quadrotors \cite{mahony2012multirotor}. Therefore, the coupling of acceleration and angular velocity makes the time-optimal trajectory of the quadrotors even more complex.

\begin{figure}[tbp]
	\centerline{\includegraphics[width=1\linewidth]{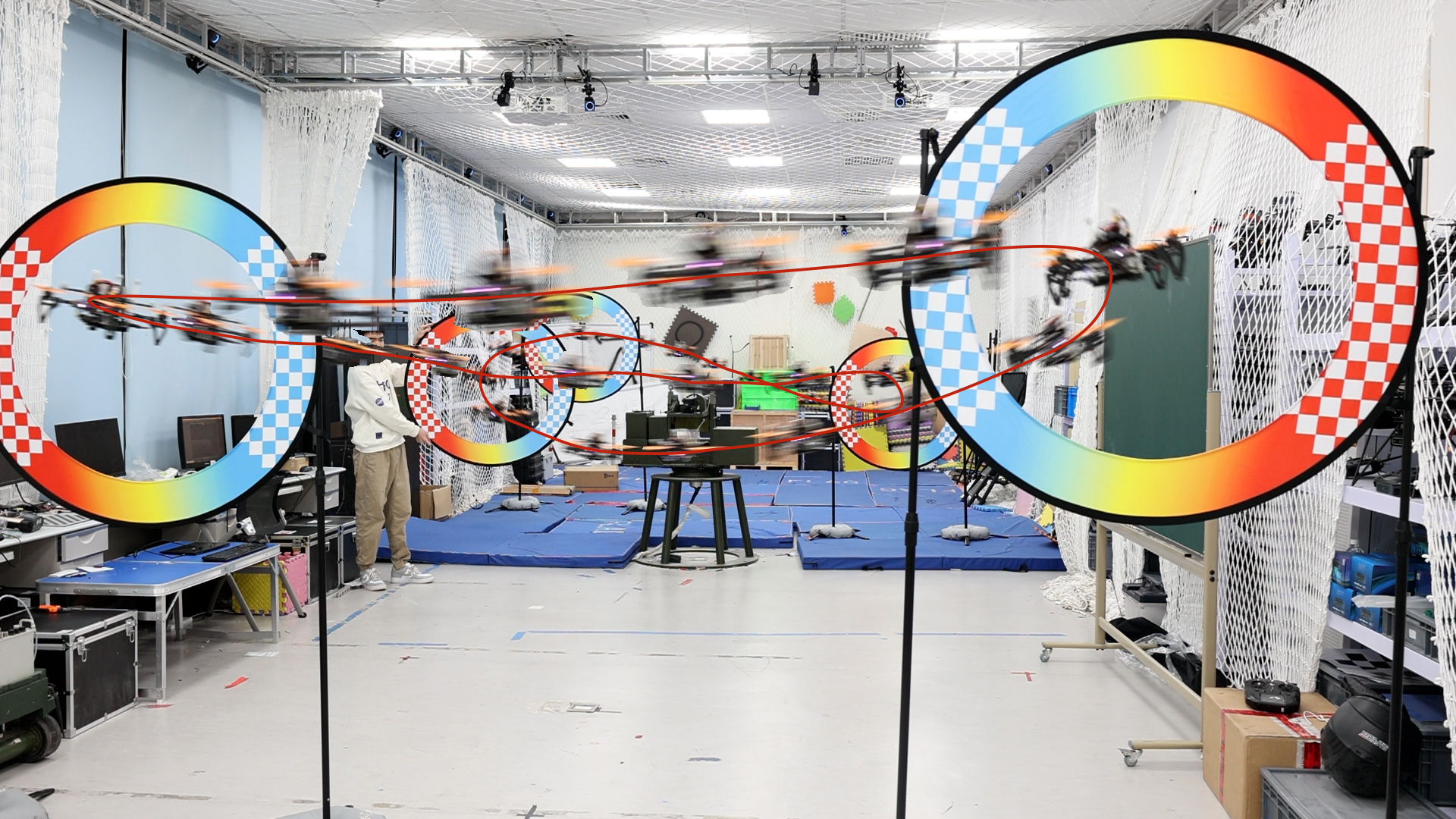}}
	\caption{Quadrotor racing through circles at a maximum speed of $10.6$ m/s. The circles have a radius of $0.8$ m, with four stationary circles and one moving circle held by a person. The time-optimal trajectory is quickly replanned when the position of the moving circle changes. Our proposed methods enable the quadrotor to efficiently replan and accurately track the updated trajectory in a fast and precise manner.}
	\label{fig1}
\end{figure}

To address this challenge, a common approach uses nonlinear numerical optimization to plan the trajectory, by discretizing it in time and treating the full quadrotor dynamics as dynamical constraints \cite{hargraves1987direct, geisert2016trajectory, zhou2020ego}. However, for multi-waypoint time-optimal flight, each waypoint must be allocated as a constraint to a specific node on the trajectory. The optimal time of passing through each waypoint is unknown, making it difficult to determine which node a waypoint should be assigned to. To tackle this issue, the work \cite{foehn2021time} introduced the so-called progress measure variables to represent the completion of a waypoint (progress) and used complementary progress constraints (CPC) to allocate the waypoint constraints, thus ensuring generation of time-optimal trajectories that satisfy the full dynamics. However, the introduction of progress measure variables and CPC renders the entire optimization problem complicated and numerically more challenging, whose solving time typically ranges from minutes to even hours. Although the waypoint constraints can be relaxed by a certain tolerance, the accuracy and optimality of generated trajectories is sacrificed.

For trajectory tracking, the two leading frameworks are the nonlinear model predictive controller (NMPC) and the differential-flatness-based controller. Due to advances in computer hardware, the NMPC approach has demonstrated superior tracking accuracy and robustness \cite{sun2022comparative}. However, when it comes to time-optimal trajectory planning, the quadrotor must operate at its motion limits, and dynamical model mismatch and unknown external disturbances can lead to trajectory tracking failure during extreme flight \cite{lavalle2006planning}, making robust trajectory tracking a challenging task. To overcome this problem, \cite{foehn2021time} proposed to generate a time-optimal trajectory with a slightly lower input upper-bound than the quadrotor's real actuator limit to ensure tracking robustness. Obviously, this approach sacrifices time-optimality.

In this letter, we propose an efficient and robust framework for time-optimal waypoint flight. The framework divides the waypoint flight mission into two layers: the time-optimal path planning under waypoint constraints and the time-adaptive trajectory tracking control. The overall framework is based on the full quadrotor dynamics and nonlinear optimization, which minimizes the flight time while maximizing the quadrotor flight capability. Our contributions  are summarized as follows.
\begin{itemize}
\item We propose a segmented time-optimal trajectory planning method with a much shorter solving time (seconds versus minutes), higher accuracy, and improved optimality in time than CPC. %The solution accuracy and optimality are both higher than CPC.
\item We propose a time-adaptive model predictive control (tMPC) method for tracking planned trajectories achieving a shorter tracking time and higher accuracy while ensuring the tracking robustness.
\item We implement the proposed time-optimal trajectory replanning method in real-world experiments with dynamic waypoints and validate its planning efficiency and tracking robustness. The source code of our system is publicly available at \url{https://github.com/BIT-KAUIS/Fast-fly}.
\end{itemize}

\section{Related Works}
\subsection{Time-optimal Trajectory Planning}
Trajectory planning has advanced significantly in simulation and experimentation over the past decade. Early research focused primarily on planning collision-free and safe trajectories. Recent studies have focused more on planning smooth, dynamically feasible, and minimum-time trajectories to enable quadrotors to fly more flexibly and quickly \cite{hehn2012performance, van2013time, spedicato2017minimum,zhang2022safety}.
Optimization-based trajectory planning methods have represented trajectories as time sequences of the quadrotor's state and control inputs, considering the objective of minimizing the flight time while complying with the quadrotor dynamics and input constraints \cite{foehn2021time}. A method by \cite{spasojevic2020perception} achieved time-minimum flight by maximizing the velocity along the given path using a path parametrization method while considering the translational and rotational dynamics of quadrotors. However, this method only optimizes velocity along the given path and does not optimize the path further.
The above trajectory planning methods did not consider the scenario where the quadrotors need to pass through given waypoints, which is common in drone racing. Time-optimal trajectory planning with waypoint constraints requires consideration of the time allocation problem, i.e., assigning waypoints to specific time steps, which is often non-trivial and hard to tackle. The recent work by \cite{foehn2021time} introduced the CPC method, considering the full-state dynamics constraints of the quadrotor and achieving truly time-optimal trajectory planning with waypoint constraints. However, the solution of CPC requires a large amount of computation, making it difficult to achieve real-time replanning.
To address this issue, \cite{romero2022time} performs real-time replanning of time-optimal trajectories based on a point-mass dynamics model and employs model predictive contouring control (MPCC) with full quadrotor dynamics \cite{romero2022model} to track the trajectory. This method has both replanning capabilities and exceeds previous methods in time optimality.

\subsection{Trajectory Tracking Control of Quadrotors}

The differential-flatness based controller (DFBC) has shown significant improvement in tracking performance during high-speed flight \cite{faessler2017differential, tal2020accurate}. Quadrotors have been proven to be differentially flat \cite{mellinger2011minimumsnap, faessler2017differential}; that is, once a time-parameterized 3D path with the heading angle is given, the attitude, angular velocity, and acceleration of quadrotors can be readily computed. These quantities can be sent to the lower-level controller to form a feedforward term, which can address the issue of model mismatches and unknown external disturbances. The work
\cite{tal2020accurate} developed the differential   control method by cascading an INDI controller after the DFBC. The INDI controller helps enhance robustness against external aerodynamic disturbances and achieves a maximum flight speed of nearly $13$ m/s and an acceleration exceeding $2$g on a real quadrotor.

Another control algorithm is MPC which predicts the future states for multiple time steps and computes the associated control inputs. Thanks to the recent advances of hardware and nonlinear optimization solvers \cite{CasadiAndersson2019,houska2011acado,Liu2021data,tac2023liuself}, NMPC based on full-state dynamics can meet real-time requirements. The work
 \cite{bicego2020nonlinear} used each rotor thrust as a control input, and  \cite{torrente2021data} used the solved internal state as the reference input which is sent to the lower-level controller. NMPC can fully exploit the capabilities of drones and account for the actuator saturation constraints. 
 
 In a recent comparative study of NMPC and DFBC by \cite{sun2022comparative}, it was shown that although DFBC is more computationally efficient and easier to implement, NMPC performs better in high-speed flight. MPC can only track trajectories that satisfy dynamic feasibility. To this end, another predictive control method, namely MPCC \cite{liniger2015optimization}, maximizes the trajectory tracking progress while considering the tracking accuracy. This method can not only track trajectories that are not dynamically feasible but also achieve approximately time-optimal flight \cite{romero2022model}.

\section{Methodology}

\subsection{Quadrotor Dynamics}\label{AA}
\begin{figure}[tbp]
    \centerline{\includegraphics[width=0.98\linewidth]{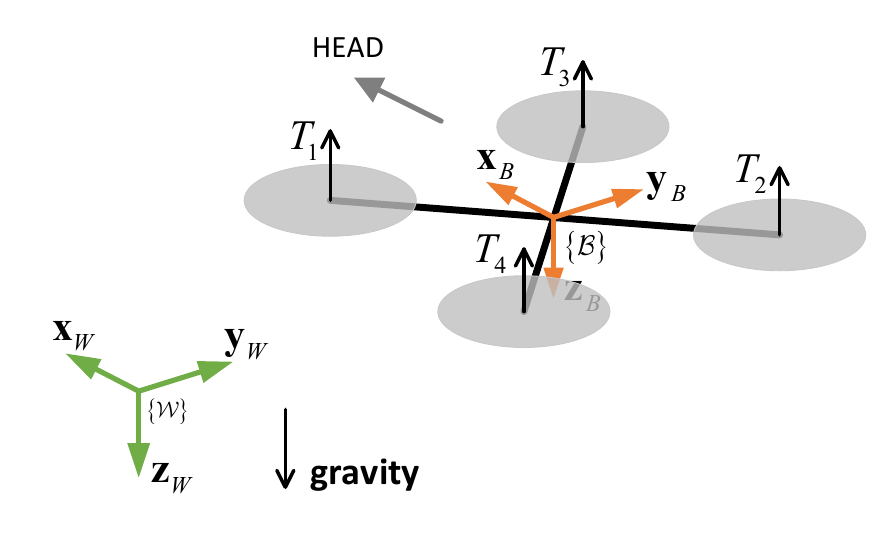}}
    \caption{A diagram showing the world frame and  body frame of a quadrotor.}
    \label{fig:frames}
\end{figure}

We represent vectors and matrices using bold lowercase and bold uppercase letters, respectively, throughout this letter. We define the world frame, denoted by $\mathcal{W}:\{{\bf x}_W, {\bf y}_W, {\bf z}_W \}$, with ${\bf z}_W$ pointing downward and aligned with gravity. The body frame, denoted by $\mathcal{B}:\{{\bf x}_B, {\bf y}_B, {\bf z}_B\}$, has ${\bf x}_B$ pointing forward and ${\bf z}_B$ pointing downward, opposite to the collective thrust. The body frame's origin is attached to the quadrotor's center of mass (CoM), as depicted in Fig.~\ref{fig:frames}.

We consider the quadrotor dynamical model used in, e.g., \cite{foehn2021time,romero2022model}, whose dynamics are given as follows
\begin{equation}
    \begin{array}{l}
        {\bf{\dot p}} = {\bf{v}}\\
        {\bf{\dot v}} = g{{\bf{z}}_W} - c{{\bf z}_B} - {\bf RD}{{\bf{R}}^\top}{\bf{v}}\\
        {\bf{\dot q}} = \frac{1}{2}{\bf{q}} \odot \left[ {\begin{array}{*{20}{c}}
        0\\
        {\boldsymbol{\omega}}
        \end{array}} \right]\\
        {\boldsymbol{\dot \omega }} = {{\bf J}^{ - 1}}\!\left( {{\boldsymbol{\tau }} - {\boldsymbol{\omega }} \times {\bf J}{\boldsymbol \omega}} \right).
  \end{array} \label{dyn}
\end{equation}
Here, ${\bf p}$ and ${\bf v}$ denote the position and velocity of the quadrotor's CoM, respectively. The symbol ${\bf q}\in\mathbb{S}\mathbb{O}(3)$ is the unit quaternion representing the rotation from $\mathcal{W}$ to $\mathcal{B}$, and ${\bf R}$ is the corresponding rotation matrix parameterized by ${\bf q}$. The symbol $\boldsymbol{\omega}$ is the angular velocity of $\mathcal{B}$ with respect to $\mathcal{W}$, $c$ and $\boldsymbol{\tau}$ are the mass-normalized collective thrust and the resultant torque generated by rotors. The symbol ${\bf D}=\mathrm{diag}(d_x,d_y,d_z)$ is the mass-normalized rotor-drag coefficients, and ${\bf J}$ is the quadrotor's inertia matrix.

For the configuration depicted in Fig.~\ref{fig:frames}, the thrust $T_i$ at each rotor $i \in \{ 1,2,3,4 \}$ can be used to decompose $c$ and ${\boldsymbol \tau}$ as follows
\begin{equation}
    c = \frac{1}{m}(T_1+T_2+T_3+T4) 
\end{equation}
\begin{equation}
    {\boldsymbol \tau} = \left[ {\begin{array}{*{20}{c}}
        {\frac{l}{{\sqrt 2 }}\left( {{T_1} + {{\rm{T}}_4} - {T_2} - {T_3}} \right)}\\
        {\frac{l}{{\sqrt 2 }}\left( {{T_1} + {T_3} - {T_2} - {T_4}} \right)}\\
        {{c_\tau }\left( {{T_3} + {T_4} - {T_1} - {T_2}} \right)}
        \end{array}} \right]
\end{equation}
where $m$ and $l$ represent the quadrotor's mass and arm length respectively. Additionally, the thrust values $T_i$ must satisfy the following constraints
\begin{equation}
    0 \leqslant T_{\min} \leqslant T_i \leqslant  T_{\max}.
\end{equation}

\subsection{Time-optimal Trajectory Planner}
The objective of a waypoint flight is, given a starting point, to guide the quadrotor through $M$ given waypoints in a specific order. In the general optimization-based planning algorithm, the time required to reach each waypoint is predetermined, and the task of passing through the waypoint is expressed as position constraints of the quadrotor at the corresponding times \cite{jorris2009three, bousson20134d}. If $\{{\bf p}_{w_i}\in\mathbb{R}^3\}_{i=1}^{N_{w}}$ denote the positions of $N_{w}$ waypoints, the optimization problem for the waypoint flight can be expressed as follows
\begin{subequations}\label{eq:cpc}
	\begin{align}
		\min_{{\bf x}_k,{\bf u}_k}\quad &J=\sum_{k=0}^{N-1}(\|{\bf x}_{k+1}\|_{\bf Q}+\|{\bf u}_k\|_{\bf R}) \label{pc_1} \\
		{\rm s.t.\,}\quad
		&\|{\bf p}_{m_i}-{\bf p}_{w_i}\|_2^2 \leqslant \delta_i^2,\quad i=1,2,\ldots,N_{w} \label{pc_2} \\
		&{\bf x}_{k+1}={\bf f}({\bf x}_{k}, {\bf u}_{k},dt) \label{pc_3} \\
		&{\bf x}_{\rm lb} \leqslant {\bf x}_{k} \leqslant {\bf x}_{\rm ub} \label{pc_4} \\
		&{\bf u}_{\rm lb} \leqslant {\bf u}_{k} \leqslant {\bf u}_{\rm ub} \label{pc_5} \\
		&{\bf x}_0={\bf x}_{\rm init} \label{pc_6}.
	\end{align}
\end{subequations}
 
\begin{figure}[tbp]
    \centerline{\includegraphics[width=\linewidth]{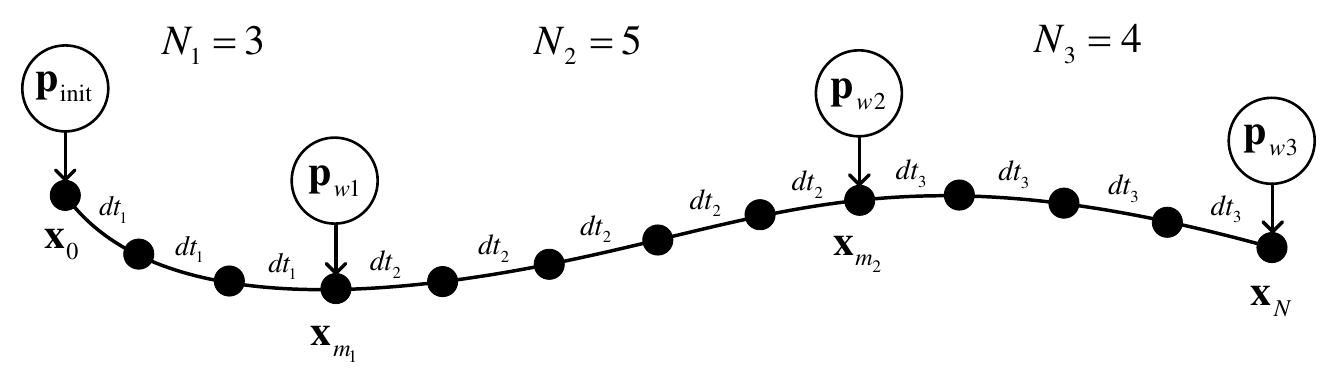}}
    \caption{Demonstration of the proposed time-optimal flight method with $N_w=3$ waypoints. The trajectory is divided into three segments, each consisting of a pre-assigned number of discrete points, with $N_1=3$, $N_2=5$, and $N_3=3$. The sampling times for the three segments are denoted by $dt_1$, $dt_2$, and $dt_3$, respectively. The three waypoint constraints are allocated to nodes $m_1=3$ and $m_2=8$, as well as the last node $N$.}
    \label{fig:topt}
\end{figure}

The objective function in \eqref{pc_1} minimizes the control cost and system energy, where $\|{\bf x}\|_{\bf A}$ is defined as ${\bf x}^\top{\bf A}{\bf x}$ for a positive definite matrix ${\bf A}={\bf A}^\top$ of suitable dimensions. The waypoint constraints are represented in \eqref{pc_2}, where $m_i\in\mathbb{N}$, ${\bf p}_{m_i}$ and $\delta_i\ge 0$ denote the time, the position of the quadrotor and the allowable position error,  when passing through the $i$-th waypoint. The dynamics constraint is represented in \eqref{pc_3}, which can be obtained from the continuous dynamics in \eqref{dyn} using the Runge-Kutta method with a sampling period of $dt>0$. The state constraints, input constraints, and initial state constraint are represented in \eqref{pc_4}--\eqref{pc_6}, respectively.

For time-optimal planning, time is treated as an optimization variable. To achieve the goal of minimizing the total flight time, the optimization objective function for time-optimal planning only considers minimizing the flight time $T$, which is not reflected in the objective of minimizing system energy and control cost in \eqref{pc_1}.

To determine $m_i$, the work of \cite{foehn2021time} introduced complementary progress constraints to optimize the values of $m_i$, which considerably increases the computation time of the resulting nonlinear optimization problem. In our method, we fix the allocation of the waypoint constraints, i.e., by pre-assigning each $m_i$ an appropriate value, and divides the trajectory into $N_{w}$ segments using the $N_w$ waypoints, each using a separate sampling interval denoted by $dt_i>0$. 
%and $dt_i$'s belonging to different segments are independent of each other. 
The number of discrete points $N_i\in\mathbb{N}$ for each trajectory segment is pre-assigned based on the distance between adjacent waypoints; see a pictorial illustration in Fig.~\ref{fig:topt}. As a result, the total flight time $T$ 
 and the allocation $m_i$'s of the waypoint constraints can be obtained as follows
\begin{subequations}
\begin{align}
	T&=\sum_{i=1}^{N_{w}}N_i \, dt_i \label{totaltime}\\
	m_i&=\sum_{j=1}^{i}N_j ,\quad \forall i=1,2,\ldots,N_w .\label{mi}
\end{align}
\end{subequations}

Furthermore, for each trajectory segment, we discretize the quadrotor dynamics using the corresponding sampling time $dt_i$ in the following form
\begin{equation}
    {\bf x}_{k_i+1}={\bf f}({\bf x}_{k_i},{\bf u}_{k_i},dt_i) \label{dynconstraint}
\end{equation}
where $k_i\in \mathbb{N}$ and $m_{i-1}\leqslant k_i\leqslant m_i$, representing the nodes (i.e., discrete points) in the $i$-th trajectory segment.

Finally, the proposed time-optimal waypoint trajectory planning problem can be summarized as follows  
\begin{subequations}\label{topt}
\begin{align}
    \min_{{\bf x}_k,{\bf u}_k,dt_i}\quad &T ~\text{in~\eqref{totaltime}}\\
  {\rm s.t.}~~\quad
    &\|{\bf p}_{m_i}-{\bf p}_{w_i}\|_2^2 \leq \delta_i^2,~\text{with $m_i$ in \eqref{mi}} \\
    &\text{Constraint in \eqref{dynconstraint}} \label{eq:dyncon} \\
    &{\bf x}_{\rm lb} \leqslant {\bf x}_{k} \leqslant {\bf x}_{\rm ub} \\
    &{\bf u}_{\rm lb} \leqslant {\bf u}_{k} \leqslant {\bf u}_{\rm ub} \\
    &{\bf x}_0={\bf x}_{\rm init}.
\end{align} 
\end{subequations}

Compared to the time-optimal flight optimization problem using CPC in \cite{foehn2021time}, our proposed Problem \eqref{topt} can be efficiently solved using interior-point methods as the allocation of each waypoint constraint is fixed. However, the nonlinear dynamics constraints \eqref{eq:dyncon} make Problem \eqref{topt} intrinsically nonconvex. Thus, to ensure the quality of the interior-point method, a good initialization is required. In order to construct such an initialization, we propose a warm-up problem, which incorporates the waypoint and dynamics constraints into the objective as penalty functions
\begin{align*}
     L_{w}&=\sum_{i=1}^{N_w}\|{\bf p}_{m_i}-{\bf p}_{w_i}\|_2^2 \\
     L_{d}&=\sum_{k=0}^{N-1}\|{\bf x}_{k+1}-{\bf f}({\bf x}_{k},{\bf u}_{k},dt_0)\|_2^2
\end{align*}
where $dt_0$ is a constant.

There exist infinitely many solutions that fulfill both the relaxed waypoint and dynamics constraints. To ensure a unique optimal solution, we introduce a regularization term for the control inputs in the objective function. This term, which is defined as $    L_{c}=\sum_{k=0}^{N-1}\|{\bf u}_k\|_{\mathbf{R}}$, considerably accelerates the convergence of the interior point method.

We can find an initial solution to Problem \eqref{topt} by properly choosing $dt_0$ and solving the warm-up problem with an interior point method, which is formulated as follows
\begin{subequations}\label{warm_up}
\begin{align}
    \min_{{\bf x}_k,{\bf u}_k}\quad &L_{w}+L_{d}+L_{c} \\
  {\rm s.t.}~~~\,
    &{\bf x}_{\rm lb} \leqslant {\bf x}_{k} \leqslant {\bf x}_{\rm ub} \\
    &{\bf u}_{\rm lb} \leqslant {\bf u}_{k} \leqslant {\bf u}_{\rm ub} \\
    &{\bf x}_0={\bf x}_{\rm init}
\end{align} 
\end{subequations}
whose solution is used as the initialization for Problem \eqref{topt}. Upon solving \eqref{topt}, we obtain the optimal sampling times $\{dt_i^*\}_{i=1}^{N_w}$ for all segments and the time-optimal trajectory points $\{{\bf x}_{k_i}^*\}_{k_i}$ between adjacent waypoints, given by
\begin{equation}
  \mathcal{T}^*\!:=\!\big\{({\bf x}_{k_i}^*, dt_i^*)|i=\!1,\ldots,N_w,\,k_i\in \mathbb{Z},\, m_{i-1} \!\leqslant k_i \!\leqslant m_i \big\} \label{topt_res}.
\end{equation}

\subsection{Time-adaptive Trajectory Tracker} \label{tracker}

The standard NMPC calculates the control commands by solving a finite-time optimal control problem with a receding horizon $H$. Specifically, the objective function is formulated as follows
\begin{subequations}\label{nmpc}
\begin{align}
    \min_{\bf u} \quad &\sum_{k=0}^{H-1}\left(\|{\bf x}_{k+1}-{\bf x}_{{\rm ref},k}\|_{\bf Q}+\|{\bf u}_k-{\bf u}_{{\rm ref},k}\|_{\bf R}\right)\\
 {\rm    s.t.}\quad
    &{\bf x}_{k+1}={\bf f}({\bf x}_{k}, {\bf u}_{k},dt) \\
    &{\bf x}_{\rm lb} \leqslant {\bf x}_{k} \leqslant {\bf x}_{\rm ub} \\
    &{\bf u}_{\rm lb} \leqslant {\bf u}_{k} \leqslant {\bf u}_{\rm ub} \\
    &{\bf x}_0={\bf x}_{\rm init}
\end{align} 
\end{subequations}
where ${\bf x}_{{\rm ref},k}$ and ${\bf u}_{{\rm ref},k}$ are the reference states and reference inputs generated from our high-level trajectory planner.

When the reference trajectory is time-optimal, it indicates that the quadrotor's performance has reached its limits. However, the nonlinearity, imprecision, and aerodynamic effect of the quadrotor's dynamical model can still affect the actual tracking error. Fast trajectory tracking using the NMPC algorithm \ref{nmpc} faces two main challenges.
\begin{enumerate}
    \item[c1)]
    The reference trajectory is typically specified as a  sequence of discrete points, and its time interval may not match that of the trajectory tracking controller. Thus, the solution obtained from time-optimal control cannot be directly used as the reference state for NMPC.
   
    \item[c2)]  Model mismatch, sensor noise, external disturbances, and delays may cause the actual tracking time progress to deviate from that obtained from high-level planning, resulting in an increased actual tracking error.
\end{enumerate}

To address the first challenge, we represent the planned trajectory \eqref{topt_res} as a time-parameterized one and use polynomials to interpolate the trajectory between discrete points. The quadrotor's differential flatness property, combined with the small-time intervals $dt_i^*$ (less than $0.1$ s in general) between discrete points ${\bf x}_{k}^*$, enables us to express the trajectory as follows
\begin{equation}
    {\bf traj}(t)=\left \{
    \begin{array}{ll}
    {\bf traj}_1(t-t_0), &t_0 \leqslant t < t_1 \\
    {\bf traj}_2(t-t_1) ,&t_1 \leqslant t < t_2 \\
  \qquad   \vdots \\
    {\bf traj}_N(t-t_{N-1}), &t_{N-1} \leqslant t < t_N  \\
    \end{array} \right . \label{traj_t}
\end{equation}
where $t_k$ represents the time associated with the state ${\bf x}_k^*$ which can be obtained by accumulating $dt_i^*$, and ${\bf traj}_k$ is a function of time and satisfies the following conditions, where ${\bf p}_k^*$ and ${\bf v}_k^*$ denote the position and velocity components of ${\bf x}_k^*$, respectively
\begin{equation}
\begin{split}
    {\bf traj}_k(0)={\bf p}_{k-1}^* \\
    {\bf traj}_k(dt_k)={\bf p}_{k}^* \\
    \frac{{\bf traj}_k(t)}{dt}\bigg |_0={\bf v}_{k-1}^* \\
    \frac{{\bf traj}_k(t)}{dt}\bigg |_{dt_k}={\bf v}_{k}^* 
\end{split}
\end{equation}
which means  the trajectory ${\bf traj}_k$ satisfies the condition of first-order continuity.

To address the second problem, we propose optimizing the initial sampling time $t_0$ of the first reference point and subsequently sampling the reference points on the trajectory every interval $dt$, denoted as 
\begin{equation}
	{\bf p}_{{\rm ref},k} = {\bf traj}\!\left(t_0+(k-1)dt\right),~ \forall k=1,2,\ldots,H.
\end{equation} 

As the trajectory planning process accounts for the quadrotor's dynamics model, the trajectory is dynamically feasible. Therefore, we can focus on tracking the reference position of the trajectory using a time-adaptive model predictive control (tMPC) problem formulation as follows:
\begin{subequations}
\begin{align}
    \min_{{\bf x}_k, {\bf u}_k, t_0}\quad &\sum_{k=1}^{H}\|{\bf p}_k-{\bf p_{ref}}_{,k}\|_2^2 \\
  {\rm  s.t.}~\;\quad 
    &{\bf x}_{k+1}={\bf f}({\bf x}_{k}, {\bf u}_{k},dt) \\
    &{\bf x}_{\rm lb} \leqslant {\bf x}_{k} \leqslant {\bf x}_{\rm ub} \\
    &{\bf u}_{\rm lb} \leqslant {\bf u}_{k} \leqslant {\bf u}_{\rm ub} \\
    &{\bf x}_0={\bf x}_{\rm init}
\end{align}
\end{subequations}
where ${\bf x}_k$ and ${\bf u}_k$ represent the state and control inputs at time step $k$, respectively, and $H$ is the prediction horizon.

\section{Experiments}

In this section, we present a set of simulations and experiments to validate the effectiveness of our proposed methods. We implemented the nonlinear optimization and optimal control problems using the open-source CasADi toolkit \cite{CasadiAndersson2019} and developed our program in the ROS environment. We designed separate ROS nodes for the trajectory planner and tracker, and implemented two nodes for simulation and actual flight, allowing for seamless switching between the two without modifying the program. The inertial and geometric parameters of the quadrotor used in our simulations and experiments are listed in Table \ref{quadconfig}.

\begin{table}[htbp]
\caption{Quadrotor Configurations}
\renewcommand\arraystretch{1.3}
\begin{center}
\begin{tabular}{ccc}
\hline
\multicolumn{2}{c}{Parameters} & Values \\
\hline
${\bf D}$            &[s$^{-1}$]   & ${\rm diag}(0.398, 0.316, 0.230)$ \\
${\bf J}$            &[gm$^2$]     & ${\rm diag}(1,2,3)$ \\
$m$                  &[kg]      & 1.2 \\
$l$                  &[m]        & 0.3 \\
$(T_{\min},T_{\max})$  &[N]        & (0, 6.9)  \\
$c_{\tau}$           &[1]        & 0.2 \\
\hline
\end{tabular} \label{quadconfig}
\end{center}
\end{table}

\subsection{Performance of Time-optimal Planner}

\emph{Comparison with the state-of-the-art methods.}
Previous work by \emph{Foehn et al.}~\cite{foehn2021time} introduced CPC to achieve time-optimal trajectory planning. This method currently benchmarks the time-optimal flight performance, having even outperformed professional human pilots in drone races. However, the time-optimal trajectory generated by the CPC method requires a significant amount of computation time, typically taking several minutes to even hours to compute the full-state time-optimal trajectory.

Our study presents a time-optimal trajectory planning approach that addresses the same problem as the CPC method, but with considerably less computation time and better solution accuracy. To evaluate the performance of our method, we employed the drone configuration provided by Foehn et al. \cite{foehn2021time}, which is summarized in Table \ref{quadconfig}, and conducted a comparative analysis with the CPC method using an equal number of discretization points for $4$ to $8$ waypoints. Our experimental results in Table~\ref{time_res} reveal that our method is orders of magnitude faster than the CPC method, while maintaining superior solution quality. Moreover, our method offers the advantage of pre-assigning waypoint constraints and the flexibility to continuously adjust the optimal time for passing through each waypoint, leading to a higher precision on the waypoint constraints in contrast to the CPC method.

\begin{table}[tbp]
\caption{Comparison with the CPC method}
\renewcommand\arraystretch{1.3}
\begin{center}
\begin{threeparttable}
    \begin{tabular}{c|cccc}
        \hline
        \multirow{2}{*}{\begin{tabular}[c]{@{}c@{}}Number of waypoints\end{tabular}} & \multicolumn{2}{c}{Optimization time [s]} & \multicolumn{2}{c}{Solution [s]} \\
            & CPC    & Ours\tnote{1}           & CPC   & Ours  \\ \hline
        $n=4$ & $417.12$ & $0.598+1.11$ & $7.973$ & $7.935$ \\
        $n=5$ & $1960$   & $1.15+1.12 $ & $8.939$ &$ 8.884 $\\
       $ n=6 $&$ 1720 $  & $1.24+1.20$  &$ 9.86 $ & $9.829$ \\
       $ n=7$ & $1980 $  & $1.33+1.92$  & $11.45 $& $11.41$ \\ 
      $  n=8$ & $2350$   &$ 1.69+2.24$  & $11.93$ & $11.65$ \\ \hline
    \end{tabular} \label{time_res}
    \begin{tablenotes}
    \footnotesize
    \item[1] 
    Total optimization time for our method, including solving time for the warm-up problem \eqref{warm_up} and the time-optimal planning problem \eqref{topt}.
    \end{tablenotes}
\end{threeparttable}
\end{center}
\end{table}

\begin{figure}[tbp]
    \centerline{\includegraphics[width=1.0\linewidth]{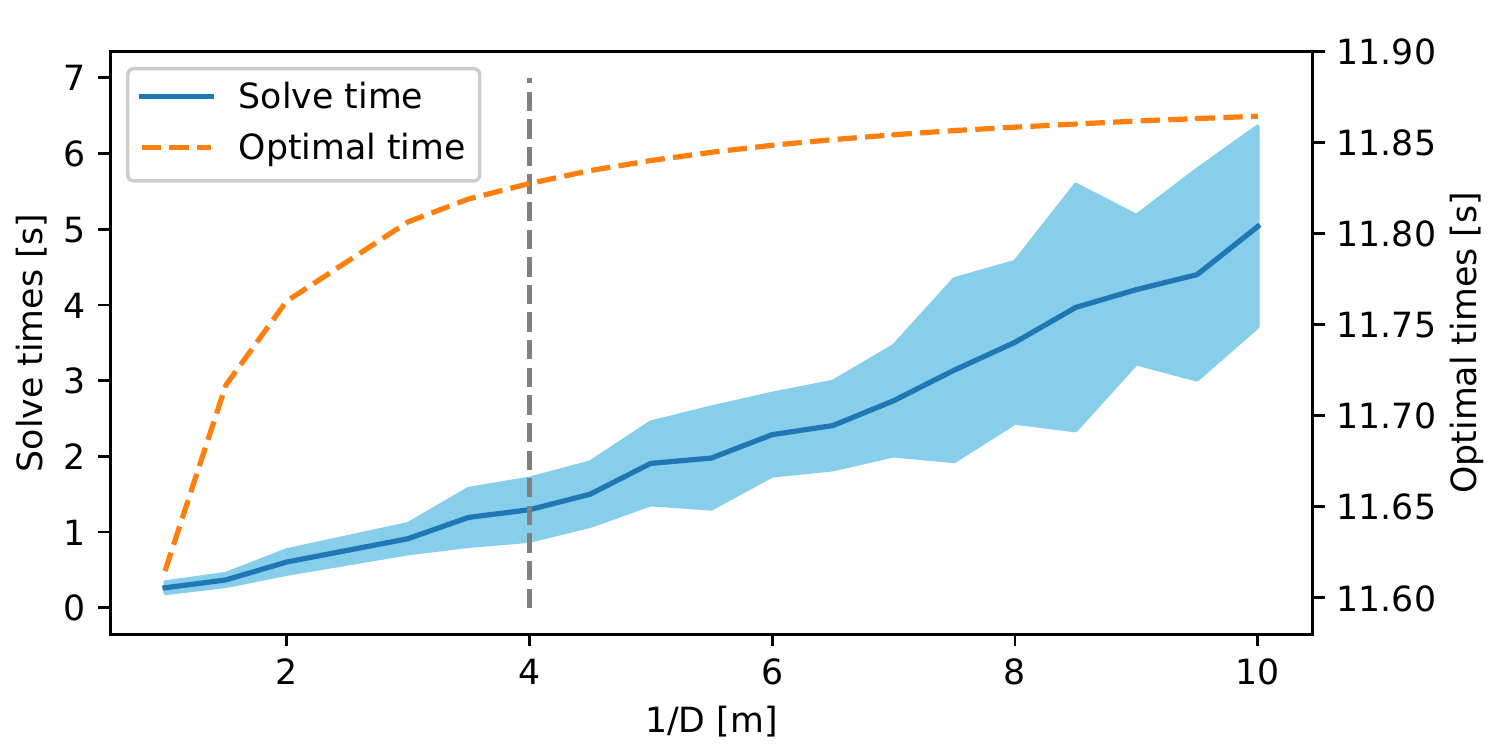}}
    \caption{
   The convergence trend of the optimal solution and the increasing trend of the solving time at different values of 1/D (average number of discrete points per meter). The location of each waypoint is fixed, and we measure the solving time $20$ times with different initial solutions. The blue line represents the mean solving time, and the blue shadow represents the standard deviation.
   }
    \label{time_analysis}
\end{figure}

\emph{Convergence analysis of the optimal solution.}
In order to improve the solution quality of our time-optimal planning method, we carefully determine the number of discretization points $N_i$ between adjacent waypoints. Despite considering the accurate drone dynamics model in the optimization problem \eqref{topt}, the finite number of nodes used for discretizing the trajectory introduces a certain level of approximation. To ensure the uniformity of each node, we determine $N_i$ based on the distance between adjacent waypoints. Specifically, we use the spatial density of nodes, denoted by $D$, to compute $N_i$ as per the following equation
\begin{equation}
    N_i=\left\lfloor \frac{\|{\bf wp}_{i}-{\bf wp}_{i-1}\|}{D}\right\rfloor 
\end{equation}
where $\lfloor \cdot  \rfloor$ represents the floor function that outputs the greatest integer less than or equal to the given value.

To yield a more accurate trajectory, we discretize the trajectory into more nodes. However, this results in increased computation time. Therefore, we investigate the convergence trend of our optimal solution and the computation time under different node densities. Fig.~\ref{time_analysis} depicts the convergence trend of the optimal solution and the computation time at various values of 1/D (average number of discrete points per meter) with a fixed location for each waypoint. We measure the solving time $20$ times with different initial solutions and calculate the mean (blue line) and standard deviation (blue shadow). As shown, with an increasing number of discrete points, the accuracy of our optimal solution gradually improves. However, the solving time increases rapidly, and the solution time becomes increasingly unstable. To balance the solution accuracy and the computation time, we choose $D=0.3$.

\subsection{Analysis of the Fast Trajectory Tracking Method}
\begin{figure}[t]    \label{track_cmp}
    \centerline{\includegraphics[width=.92\linewidth]{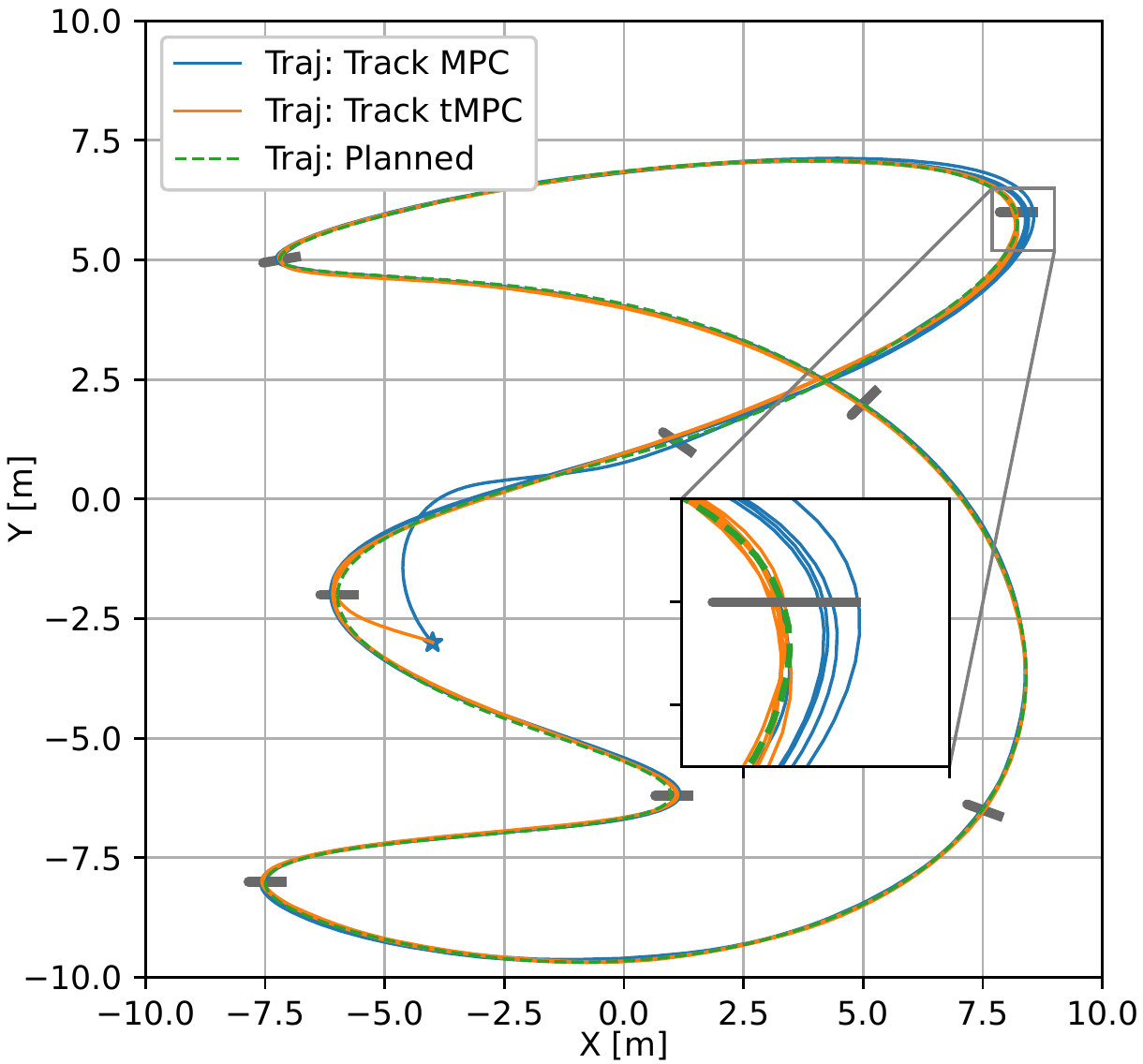}}
    \caption{Simulation comparison of the trajectory tracking performance between our tMPC and the plain-vanilla MPC. The gray line segments denote the pre-assigned waypoints, and the green dashed line represents the time-optimal trajectory. The quadrotor takes off from the blue star point and tracks the planned trajectory.
 }
\end{figure}

\begin{figure}[t]
    \centerline{\includegraphics[width=\linewidth]{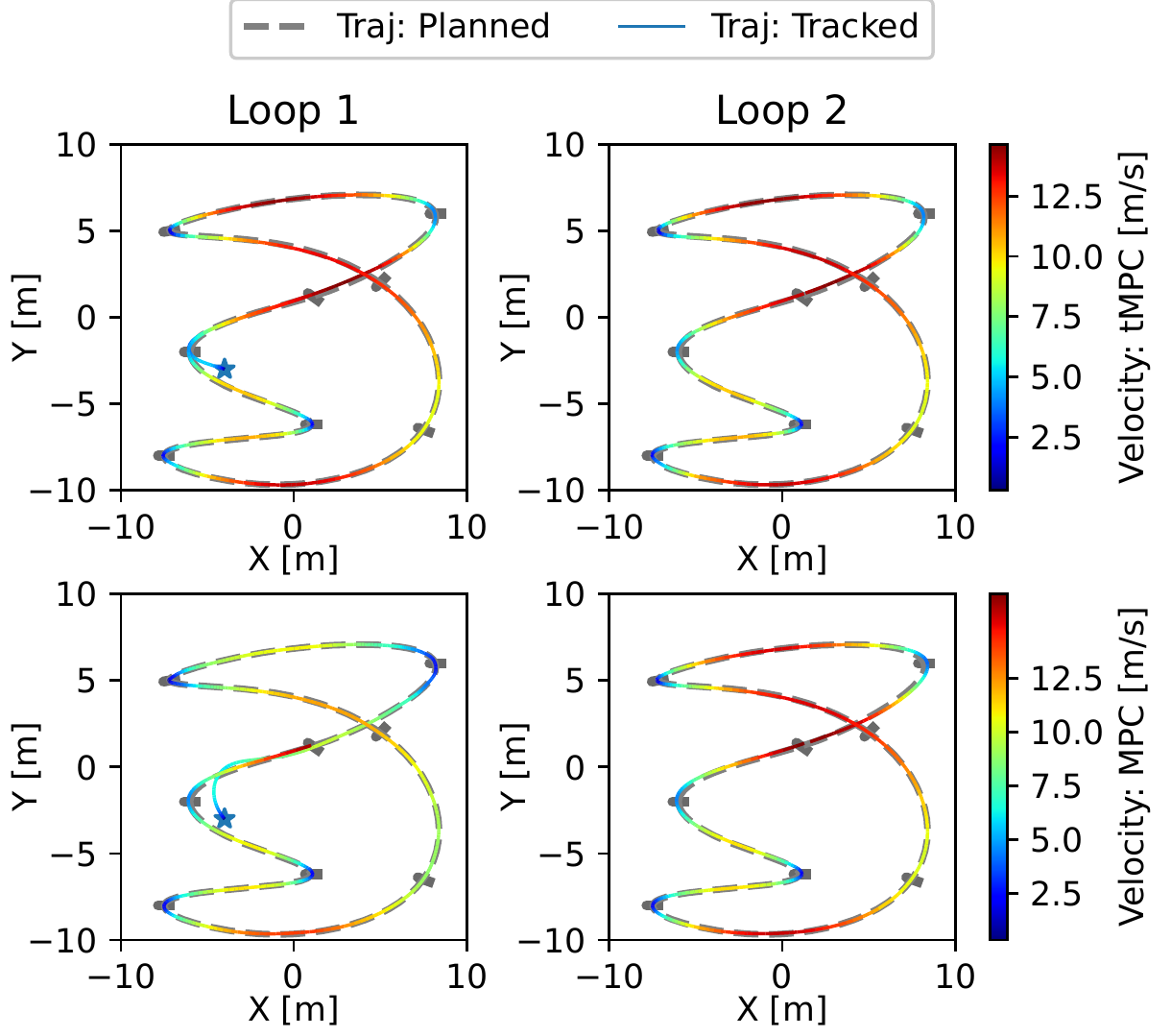}}
    \caption{Velocity tracking performance comparison between tMPC and MPC. The  two plots at the top show the velocity tracking performance of tMPC for the first and second loops after takeoff, while the two at the bottom show the velocity tracking performance of MPC for the first and second loops.}
    \label{fig:track_vel}
\end{figure}

\begin{table}[tbp]
\caption{Tracking Error under Model Mismatch}
\renewcommand\arraystretch{1.3}
\begin{center}
% \usepackage{multirow}
\begin{tabular}{l|cc|cc|cc}
    \hline
    \multirow{2}{*}{} & \multicolumn{2}{c|}{RMES [m]}   & \multicolumn{2}{c|}{Max Error [m]} & \multicolumn{2}{c}{Track Time [s]} \\ \cline{2-7} 
                      & tMPC           & MPC            & tMPC             & MPC             & tMPC             & MPC             \\ \hline
    Baseline          & $\bf{0.036}$   & $0.04$         & $\bf{0.133}$     & $0.147$         & $\bf{9.75}$      & $9.81$          \\ \hline
    $m-0.03$ kg        & $\bf{0.031}$   & $0.037$        & $\bf{0.098}$     & $0.151$         & $\bf{9.64}$      & $9.81$          \\
    $m+0.03$ kg        & $\bf{0.052}$   & $0.081$        & $\bf{0.208}$     & $0.329$         & $9.87$           & $\bf{9.83}$     \\ \hline
    $0.9 D$      & $0.044$        & $\bf{0.043}$   & $0.223$          & $\bf{0.134}$    & $\bf{9.73}$      & $9.83$          \\
    $1.1 D$      & $\bf{0.029}$   & $0.039$        & $\bf{0.092}$     & $0.148$         & $\bf{9.74}$      & $9.82$          \\ \hline
    \end{tabular}
        
\end{center} \label{track_err}
\end{table}

Upon obtaining the time-optimal trajectory from the high-level planner, we employ the trajectory tracking method in Section \ref{tracker} to control the quadrotor and track the trajectory. To assess the tracking performance, we define the root-mean-square error (RMSE) of position tracking as follows
\begin{equation}
    {\rm RMSE} = \sqrt{\frac{1}{n} \sum_{i=1}^{n} \min_t\left\|{\bf p}_i - {\bf traj}(t)\right\|^2}
\end{equation}
where ${\bf traj}(t)$ represents the time-parameterized trajectory in \eqref{traj_t}, ${\bf p}_i$ denotes the actual position of the quadrotor during the tracking process, and $n$ is the number of sampling points.

We compared our proposed trajectory tracking algorithm with the standard MPC. Our method demonstrated superior performance in terms of position tracking error, flight speed, actual flight time, and tracking robustness. When disturbances or model mismatch were present, our proposed tMPC was capable of adaptively tracking the time-optimal trajectory, whereas standard MPC failed under time-optimal flight conditions. To address this issue, we slightly slowed down the tracking progress of MPC to cope with external disturbances and actuator delays. Figure~\ref{track_cmp} shows the trajectory tracking performance of tMPC and MPC under constraints of eight waypoints. Table \ref{track_err} presents the average and maximum tracking errors and actual tracking time when the dynamic model is accurate and when there are mismatches in UAV mass or rotor drag coefficient. The results indicate that due to the inability of MPC to adapt its tracking time, the tracking accuracy of MPC varied significantly when there were model mismatches, while tMPC automatically adjusted the tracking time and maintained tracking accuracy. 
{Further, we compared the velocity tracking performance of tMPC and MPC, revealing that tMPC exhibits better tracking performance. As shown in Fig.~\ref{fig:track_vel}, tMPC can quickly track the time-optimal trajectory with desired speed for both the first and second loops after takeoff. In contrast, MPC requires a longer transition process in the first loop.}

\subsection{Real-world Waypoint Racing}

\begin{figure}[tbp]
    \centerline{\includegraphics[width=\linewidth]{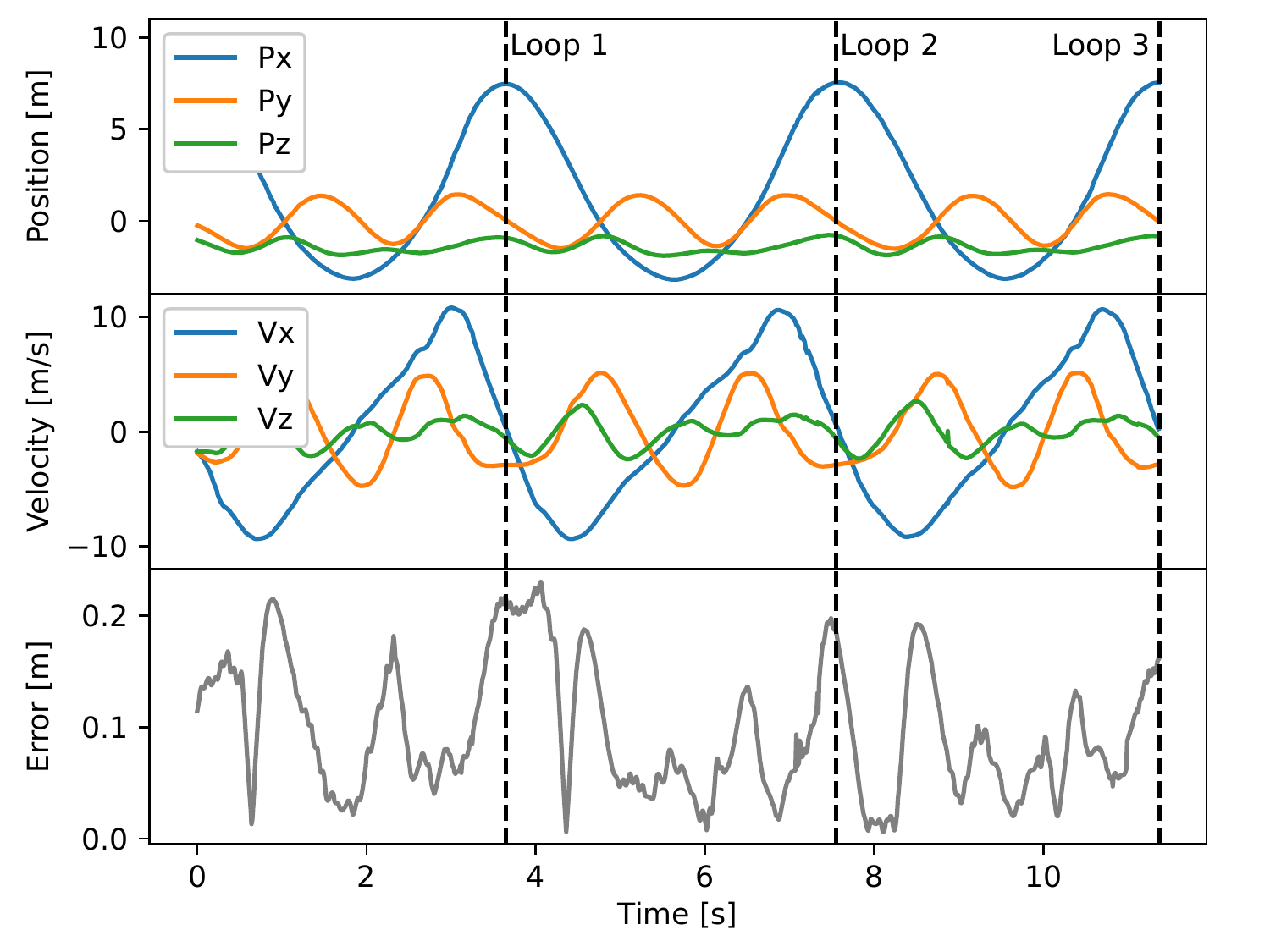}}
    \caption{Performance of time-optimal flight over static waypoints in real world. Panels from top to bottom are the position, velocity, and tracking error for three loops of actual flight. The maximum flight speed reached $10.6$ m/s, and the maximum position error during the entire trajectory tracking was $0.22$ m.}
    \label{fig:track_real_static}
\end{figure}

To further validate the performance of our algorithm, we conducted experiments on a physical quadrotor platform. %, as depicted in Fig.~\ref{fig_quadrotor}. 
The physical platform utilized the PX$4$-Vision frame and power kit and was integrated with an Intel NUC$11$ as the onboard computer. For the lower-level control, we employed the Pixhawk$4$ flight controller, which can accept angular velocity and thrust as control inputs and runs the angular velocity controller at $1,000$ Hz. In addition, the PX$4$ firmware running on the Pixhawk$4$ provided the extended Kalman filter for state estimation, which fused Inertial Measurement Unit (IMU) data with external auxiliary positioning provided by an OptiTrack motion capture system at $100$ Hz.

To demonstrate the accuracy of trajectory tracking, we conducted an experiment where the drone flew through a circle with a diameter of $0.8$ m. The experiment involved five circles, each representing a waypoint that the drone has to navigate through. 
We first conduct time-optimal flight experiments with static circles. In the experiment, we measure the position of each circle in advance and then use our proposed algorithm to plan the time-optimal trajectory and track it. We collect position and velocity data for the flight of three loops and calculate the position tracking error with respect to the time-optimal trajectory, as shown in Fig.~\ref{fig:track_real_static}. The results indicate that in the small indoor environment, we achieve a maximum flight speed of $10.6$ m/s and a tracking error of less than $0.22$ m.

\begin{figure}[tbp]
    \centerline{\includegraphics[width=0.94\linewidth]{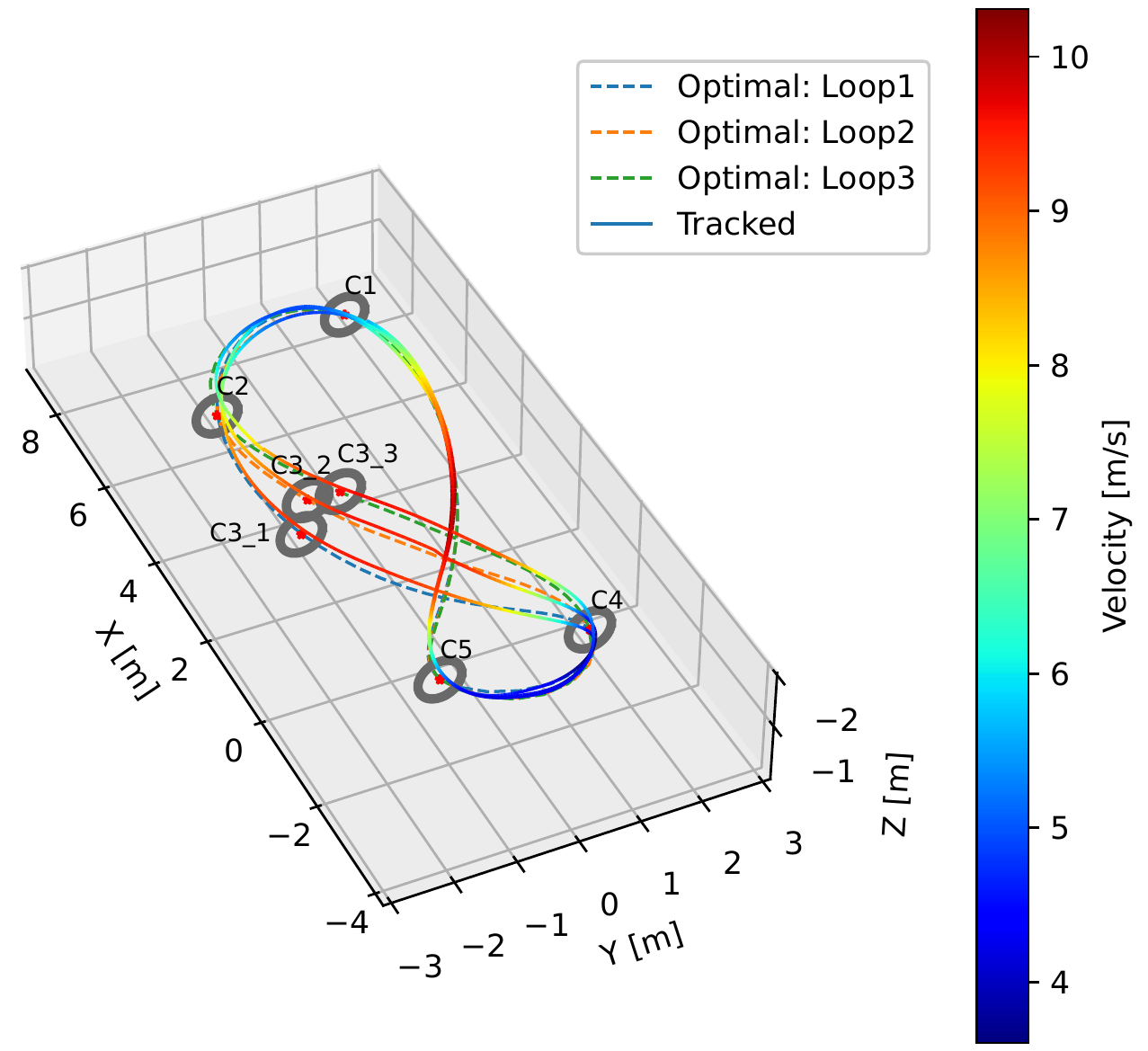}}
    \caption{Online replanning and tracking performance under dynamic waypoints in real world. In the figure, C$3\_1$, C$3\_2$, and C$3\_3$ show the positions of the third circle at different times. The dashed line represents the corresponding time-optimal trajectory. The solid lines with colors indicate the real-time position of the quadrotor, with the color signifying the flight speed.}
    \label{fig:track_real_dyn_3d}
\end{figure}

\begin{figure}[tbp]
    \centerline{\includegraphics[width=\linewidth]{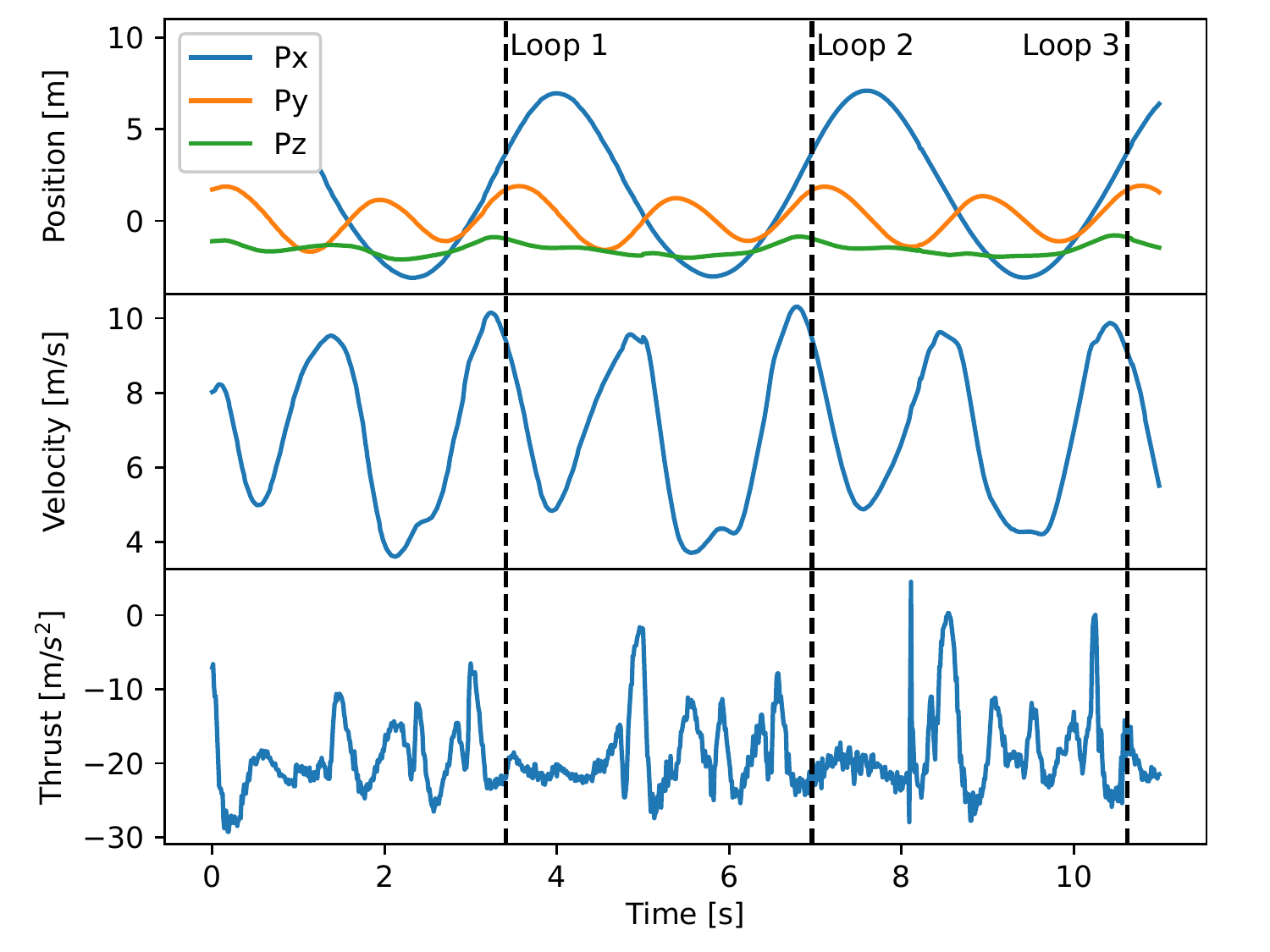}}
    \caption{Time-optimal flight data under dynamic waypoints in real world. Plots from top to bottom are the quadrotor's real-time position, velocity, and throttle. The maximum flight speed is $10.2$ m/s, and the average throttle is $21$ m/s$^2$.}
    \label{fig:track_real_dyn}
\end{figure}

To demonstrate the re-planning capability of our proposed algorithm, we conducted time-optimal flight experiments under dynamic waypoints. In the experiment, we randomly change the position of one of the waypoints and regenerate optimal trajectories in real-time, as shown in Fig.~\ref{fig:track_real_dyn_3d}. We achieved a maximum flight speed of $10.2$ m/s, with the throttle maintaining at around $21$ m/s$^2$, which is very close to the theoretical maximum throttle ($23$ m/s$^2$) of the quadrotor in the experiment, as shown in Fig.~\ref{fig:track_real_dyn}. The experimental data corroborate that our algorithm can quickly replan new time-optimal trajectories upon changing the waypoint positions (in our experiment, the replanning time was $0.12$ s) and accurately control the UAV to fly along the new trajectory.

% Furthermore, we compared the velocity tracking performance of tMPC and MPC. As shown in Fig.~\ref{fig:track_vel}, tMPC can quickly track the time-optimal trajectory with the desired speed for both the first and second loops after takeoff. In contrast, MPC requires a longer transition process in the first loop. Overall, our experimental results demonstrate the effectiveness of our algorithm in achieving high-performance trajectory tracking in real-world dynamic environments.

\section{Conclusion and Future Work}

In this letter, we have presented a novel (online) time-optimal (re)planning method for quadrotor navigation through dynamic waypoints. Our proposed tMPC algorithm effectively addressed the issue of poor robustness in time-optimal trajectory tracking. Our experimental results demonstrated that our approach can quickly plan new time-optimal trajectories and enable the quadrotor to rapidly switch to the new trajectories once the waypoint positions change. Compared to previous work, our approach achieved truly time-optimal trajectory replanning. However, we acknowledge that the proposed method does not constrain the quadrotor's yaw angle, which is often required in many applications. In future work, we will consider onboard sensor-based state estimation and gate position estimation to reduce dependence on motion capture systems. Moreover, collision-free time-optimal trajectory planning constitutes an interesting topic for future research. 
%Overall, our approach is a significant step towards efficient and robust time-optimal trajectory planning for dynamic waypoints in real-world environments.

% \begin{figure}[tbp]
%     \centerline{\includegraphics[width=\linewidth]{./figures/quadrotor.eps}}
%     \caption{Photo of the quadrotor platform}
%     \label{fig_quadrotor}
% \end{figure}

%\section*{Acknowledgment}
%The preferred spelling of the word ``acknowledgment'' in America is without 
%an ``e'' after the ``g''. Avoid the stilted expression ``one of us (R. B. 
%G.) thanks $\ldots$''. Instead, try ``R. B. G. thanks$\ldots$''. Put sponsor 
%acknowledgments in the unnumbered footnote on the first page.

\bibliographystyle{IEEEtran}
\bibliography{ref}

\begin{thebibliography}{10}
\providecommand{\url}[1]{#1}
\csname url@samestyle\endcsname
\providecommand{\newblock}{\relax}
\providecommand{\bibinfo}[2]{#2}
\providecommand{\BIBentrySTDinterwordspacing}{\spaceskip=0pt\relax}
\providecommand{\BIBentryALTinterwordstretchfactor}{4}
\providecommand{\BIBentryALTinterwordspacing}{\spaceskip=\fontdimen2\font plus
\BIBentryALTinterwordstretchfactor\fontdimen3\font minus
  \fontdimen4\font\relax}
\providecommand{\BIBforeignlanguage}[2]{{%
\expandafter\ifx\csname l@#1\endcsname\relax
\typeout{** WARNING: IEEEtran.bst: No hyphenation pattern has been}%
\typeout{** loaded for the language `#1'. Using the pattern for}%
\typeout{** the default language instead.}%
\else
\language=\csname l@#1\endcsname
\fi
#2}}
\providecommand{\BIBdecl}{\relax}
\BIBdecl

\bibitem{loianno2020special}
G.~Loianno and D.~Scaramuzza, ``Special issue on future challenges and
  opportunities in vision-based drone navigation,'' \emph{J. Field Robot.},
  vol.~37, no.~4, pp. 495--496, 2020.

\bibitem{chen2022unmanned}
J.~Chen, J.~Sun, and G.~Wang, ``From unmanned systems to autonomous intelligent
  systems,'' \emph{Eng.}, vol.~12, pp. 16--19, 2022.

\bibitem{hanover2023autonomous}
D.~Hanover, A.~Loquercio, L.~Bauersfeld, A.~Romero, R.~Penicka, Y.~Song,
  G.~Cioffi, E.~Kaufmann, and D.~Scaramuzza, ``Autonomous drone racing: {A}
  survey,'' \emph{arXiv preprint arXiv:2301.01755}, 2023.

\bibitem{foehn2021time}
P.~Foehn, A.~Romero, and D.~Scaramuzza, ``Time-optimal planning for quadrotor
  waypoint flight,'' \emph{Sci. Robot.}, vol.~6, no.~56, p. eabh1221, 2021.

\bibitem{romero2022time}
A.~Romero, R.~Penicka, and D.~Scaramuzza, ``Time-optimal online replanning for
  agile quadrotor flight,'' \emph{IEEE Robot. Autom. Lett.}, vol.~7, no.~3, pp.
  7730--7737, 2022.

\bibitem{mahony2012multirotor}
R.~Mahony, V.~Kumar, and P.~Corke, ``Multirotor aerial vehicles: {M}deling,
  estimation, and control of quadrotor,'' \emph{IEEE Robot. Autom. Mag.},
  vol.~19, no.~3, pp. 20--32, 2012.

\bibitem{hargraves1987direct}
C.~R. Hargraves and S.~W. Paris, ``Direct trajectory optimization using
  nonlinear programming and collocation,'' \emph{J. Guid. Control Dyn.},
  vol.~10, no.~4, pp. 338--342, 1987.

\bibitem{geisert2016trajectory}
M.~Geisert and N.~Mansard, ``Trajectory generation for quadrotor based systems
  using numerical optimal control,'' in \emph{IEEE Int. Conf. Robot.
  Autom.}\hskip 1em plus 0.5em minus 0.4em\relax IEEE, 2016, pp. 2958--2964.

\bibitem{zhou2020ego}
X.~Zhou, Z.~Wang, H.~Ye, C.~Xu, and F.~Gao, ``{EGO-planner: An ESDF}
  gradient-based local planner for quadrotors,'' \emph{IEEE Robot. Autom.
  Lett.}, vol.~6, no.~2, pp. 478--485, 2021.

\bibitem{sun2022comparative}
S.~Sun, A.~Romero, P.~Foehn, E.~Kaufmann, and D.~Scaramuzza, ``A comparative
  study of nonlinear {MPC} and differential-flatness-based control for
  quadrotor agile flight,'' \emph{IEEE Trans. Robot.}, vol.~38, no.~6, pp.
  3357--3373, 2022.

\bibitem{lavalle2006planning}
S.~M. LaValle, \emph{{Planning Algorithms}}.\hskip 1em plus 0.5em minus
  0.4em\relax Cambridge University Press, 2006.

\bibitem{hehn2012performance}
M.~Hehn, R.~Ritz, and R.~D'Andrea, ``Performance benchmarking of quadrotor
  systems using time-optimal control,'' \emph{Autonomous Robots}, vol.~33, pp.
  69--88, 2012.

\bibitem{van2013time}
W.~Van~Loock, G.~Pipeleers, and J.~Swevers, ``Time-optimal quadrotor flight,''
  in \emph{2013 European Control Conference}.\hskip 1em plus 0.5em minus
  0.4em\relax IEEE, 2013, pp. 1788--1792.

\bibitem{spedicato2017minimum}
S.~Spedicato and G.~Notarstefano, ``Minimum-time trajectory generation for
  quadrotors in constrained environments,'' \emph{IEEE Trans. Control Syst.
  Technol.}, vol.~26, no.~4, pp. 1335--1344, 2017.

\bibitem{zhang2022safety}
W.~Zhang, J.~Jia, S.~Zhou, K.~Guo, X.~Yu, and Y.~Zhang, ``A safety planning and
  control architecture applied to a quadrotor autopilot,'' \emph{IEEE Robot.
  Autom. Lett.}, vol.~8, no.~2, pp. 680--687, 2022.

\bibitem{spasojevic2020perception}
I.~Spasojevic, V.~Murali, and S.~Karaman, ``Perception-aware time optimal path
  parameterization for quadrotors,'' in \emph{IEEE Int. Conf. Robot.
  Autom.}\hskip 1em plus 0.5em minus 0.4em\relax IEEE, 2020, pp. 3213--3219.

\bibitem{romero2022model}
A.~Romero, S.~Sun, P.~Foehn, and D.~Scaramuzza, ``Model predictive contouring
  control for time-optimal quadrotor flight,'' \emph{IEEE Trans. Robot.},
  vol.~38, no.~6, pp. 3340--3356, 2022.

\bibitem{faessler2017differential}
M.~Faessler, A.~Franchi, and D.~Scaramuzza, ``Differential flatness of
  quadrotor dynamics subject to rotor drag for accurate tracking of high-speed
  trajectories,'' \emph{IEEE Robot. Autom. Lett.}, vol.~3, no.~2, pp. 620--626,
  2017.

\bibitem{tal2020accurate}
E.~Tal and S.~Karaman, ``Accurate tracking of aggressive quadrotor trajectories
  using incremental nonlinear dynamic inversion and differential flatness,''
  \emph{IEEE Trans. Control Syst. Technol.}, vol.~29, no.~3, pp. 1203--1218,
  2020.

\bibitem{mellinger2011minimumsnap}
D.~Mellinger and V.~Kumar, ``Minimum snap trajectory generation and control for
  quadrotors,'' in \emph{IEEE Int. Conf. Robot. Autom.}\hskip 1em plus 0.5em
  minus 0.4em\relax IEEE, 2011, pp. 2520--2525.

\bibitem{CasadiAndersson2019}
J.~A.~E. Andersson, J.~Gillis, G.~Horn, J.~B. Rawlings, and M.~Diehl,
  ``{CasADi} -- {A} software framework for nonlinear optimization and optimal
  control,'' \emph{Mathematical Programming Computation}, vol.~11, no.~1, pp.
  1--36, 2019.

\bibitem{houska2011acado}
B.~Houska, H.~J. Ferreau, and M.~Diehl, ``{ACADO toolkit—An} open-source
  framework for automatic control and dynamic optimization,'' \emph{Optim.
  Control Appl. Methods}, vol.~32, no.~3, pp. 298--312, 2011.

\bibitem{Liu2021data}
W.~Liu, J.~Sun, G.~Wang, F.~Bullo, and J.~Chen, ``Data-driven resilient
  predictive control under {Denial-of-Service},'' \emph{IEEE Trans. Autom.
  Control}, pp. 1--16, Sept. 2022, doi: 10.1109/TAC.2022.3209399.

\bibitem{tac2023liuself}
------, ``Data-driven self-triggered control via trajectory prediction,''
  \emph{IEEE Trans. Autom. Control}, pp. 1--8, 2023,
  DOI:10.1109/TAC.2023.3244116.

\bibitem{bicego2020nonlinear}
D.~Bicego, J.~Mazzetto, R.~Carli, M.~Farina, and A.~Franchi, ``Nonlinear model
  predictive control with enhanced actuator model for multi-rotor aerial
  vehicles with generic designs,'' \emph{J. Intell. Robot. Syst.}, vol. 100,
  pp. 1213--1247, 2020.

\bibitem{torrente2021data}
G.~Torrente, E.~Kaufmann, P.~F{\"o}hn, and D.~Scaramuzza, ``Data-driven {MPC}
  for quadrotors,'' \emph{IEEE Robot. Autom. Lett.}, vol.~6, no.~2, pp.
  3769--3776, 2021.

\bibitem{liniger2015optimization}
A.~Liniger, A.~Domahidi, and M.~Morari, ``Optimization-based autonomous racing
  of 1: 43 scale {RC} cars,'' \emph{Optim. Control Appl. Methods}, vol.~36,
  no.~5, pp. 628--647, 2015.

\bibitem{jorris2009three}
T.~R. Jorris and R.~G. Cobb, ``Three-dimensional trajectory optimization
  satisfying waypoint and no-fly zone constraints,'' \emph{J. Guid. Control
  Dyn.}, vol.~32, no.~2, pp. 551--572, 2009.

\bibitem{bousson20134d}
K.~Bousson and P.~F. Machado, ``{4D} trajectory generation and tracking for
  waypoint-based aerial navigation,'' \emph{Trans. Syst. Control}, no.~3, pp.
  105--119, 2013.

\end{thebibliography}
\end{document}